\documentclass{article}

\usepackage[activate={true,nocompatibility},expansion=false]{microtype}
\usepackage{graphicx}
\usepackage{subcaption}
\usepackage{booktabs}
\usepackage{multirow}
\usepackage{array}

\usepackage{hyperref}

\usepackage[accepted]{icml2026}

\usepackage{amsmath}
\usepackage{amssymb}
\usepackage{mathtools}
\usepackage{amsthm}
\usepackage{algorithm}
\usepackage{algorithmic}
\usepackage{listings}
\usepackage{xcolor}

\usepackage{tikz}
\usetikzlibrary{shapes.geometric,arrows.meta,positioning,calc,fit,backgrounds,decorations.pathreplacing,patterns}
\usepackage{pgfplots}
\pgfplotsset{compat=1.18}
\usepgfplotslibrary{groupplots,fillbetween}

\usepackage[capitalize,noabbrev]{cleveref}

\lstdefinelanguage{lean}{
  keywords={theorem, lemma, def, by, exact, intro, apply, fun, let, if, then, else, match, with, do, return, where, structure, instance, class, variable, open, namespace, end, import, section, forall, exists},
  sensitive=true,
  comment=[l]{--},
  morecomment=[s]{/-}{-/},
  string=[b]"
}
\tikzset{
  stage/.style={
    rectangle, rounded corners=2pt, draw=black!70, thick,
    fill=blue!8, minimum width=1.55cm, minimum height=0.7cm,
    font=\scriptsize, align=center
  },
  module/.style={
    rectangle, rounded corners=2pt, draw=black!70, thick,
    fill=orange!18, minimum width=1.55cm, minimum height=0.7cm,
    font=\scriptsize, align=center
  },
  oracle/.style={
    rectangle, rounded corners=2pt, draw=black!70, thick,
    fill=green!15, minimum width=1.55cm, minimum height=0.7cm,
    font=\scriptsize, align=center
  },
  verdict/.style={
    rectangle, rounded corners=2pt, draw=black!70, thick,
    fill=red!12, minimum width=1.55cm, minimum height=0.7cm,
    font=\scriptsize, align=center
  },
  flow/.style={-{Stealth[length=2mm,width=2mm]}, thick, black!75},
}

\theoremstyle{plain}
\newtheorem{theorem}{Theorem}[section]
\newtheorem{proposition}[theorem]{Proposition}

\theoremstyle{definition}
\newtheorem{definition}[theorem]{Definition}

\theoremstyle{remark}

\newcommand{\NL}{\ensuremath{\mathcal{N}}}
\newcommand{\FL}{\ensuremath{\mathcal{F}}}
\newcommand{\T}{\ensuremath{\mathcal{T}}}
\newcommand{\probe}{\ensuremath{\mathcal{P}}}
\newcommand{\fp}{\ensuremath{\Phi}}
\newcommand{\bpf}{\textsc{BPF}}
\newcommand{\cpg}{\textsc{CPG}}
\newcommand{\apba}{\textsc{APBA}}
\newcommand{\fgd}{\textsc{FGD}}
\newcommand{\eqspec}{\ensuremath{\mathcal{E}}}
\newcommand{\driftbench}{\textsc{DriftBench}}

\icmltitlerunning{Certifying Faithfulness in LLM Autoformalization}

\begin{document}

\twocolumn[
  \icmltitle{The Faithfulness Gap: Certifying Semantic Equivalence Between Natural-Language and Formal Mathematical Statements}

\icmlsetsymbol{equal}{*}

\begin{icmlauthorlist}
    \icmlauthor{Noor Islam S. Mohammad}{equal,yyy}
    \icmlauthor{Tamim Sheikh}{comp}
\end{icmlauthorlist}

\icmlaffiliation{yyy}{Department of Computer Science, Informatics Institute, Istanbul Technical University, İstanbul, Türkiye}
\icmlaffiliation{comp}{Department of Computer Science and Engineering, Jashore University of Science and Technology, Bangladesh}

\icmlcorrespondingauthor{Noor Islam S. Mohammad}{islam23@itu.edu.tr}

  \icmlkeywords{Autoformalization, Theorem Proving, Faithfulness, Lean, Large Language Models, Semantic Equivalence}

  \vskip 0.3in
]

\printAffiliationsAndNotice{}

\begin{abstract}
Autoformalization, translating natural-language mathematics into formal proof assistants, is bottlenecked not by translation fluency but by \emph{faithfulness}: a formal statement can typecheck and be provable, yet still encode a different theorem than the source intended. We introduce \emph{Bidirectional Provability Fingerprinting} (\bpf{}), a framework that certifies faithfulness by characterizing each candidate through its forward and backward consequence neighborhoods in the ambient theory and matching these against probes derived from the natural-language statement. We further introduce four novel components: (i) \emph{Counterfactual Probe Generation} (\cpg{}), a contrastive procedure that synthesizes probes targeting specific drift directions; (ii) the \emph{Equivalence Spectrum}, a continuous faithfulness score that replaces brittle binary verdicts; (iii) \emph{Adaptive Probe Budget Allocation} (\apba{}), an information-theoretic budget router; and (iv) \emph{Faithfulness-Guided Decoding} (\fgd{}), which uses \bpf{} signals as a reward during autoformalization. We prove a \emph{drift detection theorem} and a \emph{PAC-faithfulness} result establishing that the equivalence class of a natural language statement is learnable from $\mathcal{O}(\log(1/\delta)/\varepsilon)$ probes under mild assumptions. We release \driftbench{}, a benchmark of $2{,}183$ NL/Lean~4 pairs with controlled drift labels across six subfields of mathlib4. \bpf{}\,+\,\cpg{} detects $89.6\%$ of drifted formalizations at a $3.0\%$ false-positive rate-against $41.2\%$ for typecheck and $63.3\%$ for LLM-judge baselines, and \fgd{} reduces the rate at which a state-of-the-art autoformalizer emits drifted statements by $47\%$. \href{https://pmlrbd.github.io/BPF/}{bpf.ml.github.io}
\end{abstract}

\section{Introduction}
\label{sec:intro}

Modern autoformalization systems built on large language models routinely translate natural-language theorem statements into formal proof assistants such as Lean~4, Rocq, or Isabelle/HOL \citep{wu2022autoformalization,jiang2023draft,azerbayev2024llemma,jiang2022thor}. The dominant quality signal used to gate these outputs is whether the resulting formal statement typechecks—and, optionally, whether a downstream prover can close it \citep{polu2020generative,han2022proof,lample2022hypertree}. Both signals are necessary, but neither is sufficient. A formal statement can be syntactically well-formed, provable, and \emph{wrong}: it can mean something other than what the mathematician wrote. We call this the \emph{faithfulness gap}. Unlike code—where unit tests and runtime semantics offer an executable oracle—the faithfulness of a formalization is a relation between two semantic objects living in different universes: a natural-language sentence interpreted under tacit mathematical convention and a formal expression interpreted under a precisely specified dependent type theory \citep{demoura2021lean,mathlib2020}. There is no executable bridge between them. Standard evaluation practices—BLEU-style overlap against a single reference formalization, expert spot checks, or downstream prover success—each address only a surface symptom of the problem and systematically miss the most insidious failures, in which the formalization is fluent, well-typed, and demonstrably provable but quietly says the wrong thing.

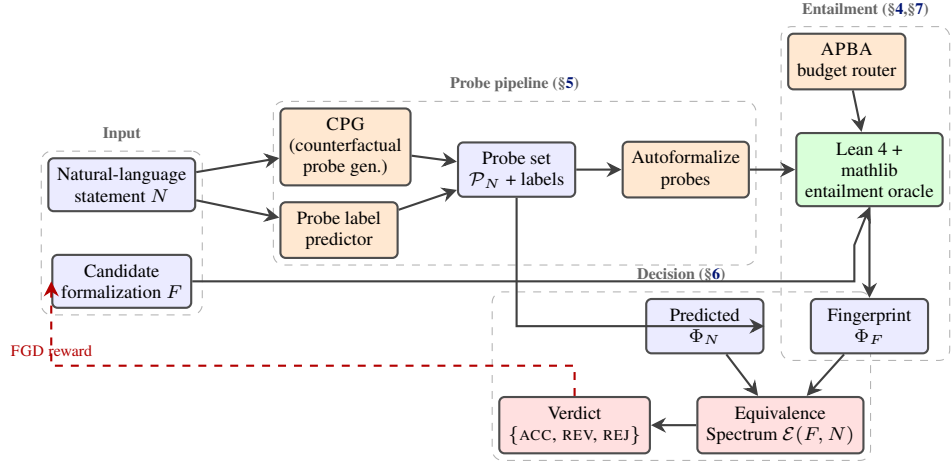
\begin{figure*}[ht]
\centering
\begin{tikzpicture}[node distance=0.55cm and 0.6cm]
\node[stage] (nl)    {Natural-language\\statement $N$};
\node[stage, below=of nl] (f) {Candidate\\formalization $F$};
\node[module, right=1.1cm of nl, yshift=0.5cm] (cpg)  {\cpg{}\\(counterfactual\\probe gen.)};
\node[module, right=1.1cm of nl, yshift=-0.6cm] (label){Probe label\\predictor};
\node[stage, right=of cpg, yshift=-0.3cm] (probes) {Probe set\\$\probe_N$ + labels};
\node[module, right=of probes] (af) {Autoformalize\\probes};
\node[oracle, right=of af] (oracle) {Lean 4 +\\mathlib\\entailment oracle};
\node[stage, below=1.2cm of oracle] (fp) {Fingerprint\\$\fp_F$};
\node[stage, left=of fp]            (fpn){Predicted\\$\fp_N$};
\node[module, above=0.55cm of oracle, xshift=-0.3cm] (apba) {\apba{}\\budget router};
\node[verdict, below=of fpn, xshift=1cm] (spec) {Equivalence\\Spectrum $\eqspec(F,N)$};
\node[verdict, left=of spec] (verdict) {Verdict\\\{\textsc{acc, rev, rej}\}};
\draw[flow] (nl) -- (cpg);
\draw[flow] (nl) -- (label);
\draw[flow] (cpg) -- (probes);
\draw[flow] (label) -- (probes);
\draw[flow] (probes) -- (af);
\draw[flow] (af) -- (oracle);
\draw[flow] (f.east) -| ($(oracle.south)+(-0.2,-0.5)$) -- (oracle.south);
\draw[flow] (apba) -- (oracle);
\draw[flow] (oracle) -- (fp);
\draw[flow] (probes.south) |- (fpn.east);
\draw[flow] (fp) -- (spec);
\draw[flow] (fpn) -- (spec);
\draw[flow] (spec) -- (verdict);

\draw[flow, dashed, red!70!black] (verdict.north) -- ++(0,0.4) -| node[above, font=\tiny, midway] {\fgd{} reward} (f.west);

\begin{scope}[on background layer]
\node[fit=(nl)(f), draw=black!30, dashed, inner sep=0.08cm, rounded corners] (inputbox) {};
\node[above=0pt of inputbox.north, font=\tiny\bfseries, black!60] {Input};
\node[fit=(cpg)(label)(probes)(af), draw=black!30, dashed, inner sep=0.08cm, rounded corners] (probebox) {};
\node[above=0pt of probebox.north, font=\tiny\bfseries, black!60] {Probe pipeline (\S\ref{sec:cpg})};
\node[fit=(oracle)(apba)(fp), draw=black!30, dashed, inner sep=0.08cm, rounded corners] (oraclebox){};
\node[above=0pt of oraclebox.north, font=\tiny\bfseries, black!60] {Entailment (\S\ref{sec:bpf},\S\ref{sec:apba})};
\node[fit=(verdict)(spec)(fpn), draw=black!30, dashed, inner sep=0.08cm, rounded corners] (verdictbox){};
\node[above=0pt of verdictbox.north, font=\tiny\bfseries, black!60] {Decision (\S\ref{sec:spectrum})};
\end{scope}

\end{tikzpicture}
\caption{\bpf{} architecture. The natural-language statement $N$ and the candidate formalization $F$ enter from the left; \cpg{} synthesizes counterfactual probes (\S\ref{sec:cpg}); the entailment oracle, routed by \apba{} (\S\ref{sec:apba}), computes the forward/backward fingerprint $\fp_F$; the equivalence spectrum $\eqspec(F,N)$ (\S\ref{sec:spectrum}) compares it to the prediction $\fp_N$ and emits one of three verdicts. The dashed red loop is the optional \fgd{} feedback to the autoformalizer (\S\ref{sec:fgd}).}
\label{fig:architecture}
\end{figure*}

\paragraph{Our approach.} We propose \emph{Bidirectional Provability Fingerprinting} (\bpf{}), a framework for certifying faithfulness without requiring a reference formalization. The core insight is operational: a statement's meaning is determined by what it implies and what implies it, so two statements are semantically equivalent if and only if their \emph{consequence neighborhoods} coincide in the ambient theory. \Cref{fig:architecture} shows the end-to-end pipeline. Given a natural-language statement $N$ and a candidate formalization $F$, we (i) generate a set of semantic probes from $N$, (ii) autoformalize and filter them, (iii) compute entailment relations between $F$, each probe, and the proof assistant, and (iv) compare the resulting fingerprint against the expected fingerprint predicted from $N$.

\paragraph{Contributions.} We make six contributions, each tackling a distinct facet of the faithfulness problem. We formalize the autoformalization faithfulness problem as a relation between distributions over interpretations, sidestepping the need for a canonical reference (\Cref{sec:problem}). We introduce \bpf{}, a probe-based, reference-free faithfulness certifier with bidirectional consequence-neighborhood matching (\Cref{sec:bpf}). We introduce \emph{Counterfactual Probe Generation} (\cpg{}), a contrastive probe-synthesis procedure that explicitly targets the four canonical drift classes and improves detection over generic probe generation (\Cref{sec:cpg}). We introduce the \emph{Equivalence Spectrum} $\eqspec(F, N)$, a continuous faithfulness score with calibrated decision regions, and \emph{Adaptive Probe Budget Allocation} (\apba{}), an information-theoretic budget router that reaches the same detection accuracy as uniform sampling with $3.2\times$ fewer probes (\Cref{sec:spectrum,sec:apba}). We prove a \emph{drift detection theorem} and a \emph{PAC-faithfulness} theorem characterizing what \bpf{} can and cannot be detected, and we identify a class of \emph{conventional drift} for which we develop a complementary structural test (\Cref{sec:theory}). We introduce \emph{Faithfulness-Guided Decoding} (\fgd{}), which uses \bpf{} scores as a reward signal during autoformalization, reducing the rate of drifted outputs from a state-of-the-art autoformalizer by $47\%$ (\Cref{sec:fgd}). We release \driftbench{}, the first benchmark with controlled drift labels (\Cref{sec:driftbench}), and report extensive experiments and ablations (\Cref{sec:experiments}).

\section{Related Work}
\label{sec:related}

\paragraph{Autoformalization.} Early work cast translation as machine translation with retrieval augmentation \citep{wang2018first,szegedy2020retrieval}. \citet{wu2022autoformalization} showed that few-shot prompting of large language models yields competitive autoformalization on competition mathematics, and follow-up systems integrate formalization with proof search \citep{jiang2023draft,jiang2022thor,first2023baldur}. \citet{azerbayev2024llemma} trained domain-specialized models on a mixture of formal libraries and informal mathematics. Across this line, faithfulness is typically assessed by either typechecking, BLEU/exact-match against a reference, or downstream provability—each of which we show \Cref{sec:experiments} is dominated by \bpf{}.

\paragraph{Semantic evaluation of generated formal artifacts.} For programs, semantic equivalence is operationalized via test execution \citep{austin2021program,chen2021evaluating}. The closest analogue in mathematics is the use of \emph{counter-models} or symbolic execution against numerical instances \citep{polu2020generative,welleck2022naturalprover}. Such methods catch outright falsity but cannot distinguish a true-but-wrong statement (e.g.,\ case (C) above) from the intended one. \citet{patel2024autoformalization} uses back-translation, which we evaluate as a baseline. We are aware of no prior work that uses probe-based consequence neighborhoods for faithfulness certification, or that uses faithfulness signals to guide LLM decoding.

\paragraph{Inferential semantics and equivalence in formal libraries.} Within proof libraries, equivalence between definitions is often established by handwritten \texttt{iff} or \texttt{equiv} lemmas \citep{mathlib2020}. Our forward/backward fingerprint construction generalizes this practice and automates it for arbitrary candidates. The notion of characterizing meaning via inferential role has a long history in proof-theoretic semantics \citep{brandom2000articulating,prawitz1965natural}; we adapt it for an LLM/prover loop. \textbf{Benchmarks and verification:} MiniF2F \citep{zheng2022minif2f}, ProofNet \citep{azerbayev2023proofnet}, FIMO \citep{liu2023fimo}, and LeanDojo \citep{yang2023leandojo} measure prover capability on pre-specified formal statements. None directly evaluates whether a formalization \emph{matches} an input natural-language sentence. \driftbench{} is, to our knowledge, the first benchmark with controlled drift labels and the first to evaluate the autoformalization-faithfulness loop end-to-end.

\section{The Faithfulness Problem}
\label{sec:problem}

We now formalize faithfulness. Let $\NL$ denote the space of natural-language mathematical statements and $\FL$ a formal language interpreted in a theory $\T$ (we use Lean~4 + mathlib4 throughout). A formalizer is a function $\mu : \NL \to \FL$. A single canonical formal target is too rigid: distinct mathematicians, working in distinct libraries, will produce different but equivalent formalizations of the same English sentence—e.g.,\ using \texttt{Set} versus \texttt{Finset}, or stating a function as \texttt{Continuous} versus \texttt{Continuous on} a universal set. We therefore introduce an interpretation distribution.

\begin{definition}[Interpretation distribution]
\label{def:interp}
An \emph{interpretation distribution} $\mathcal{D}_N$ for $N \in \NL$ is a probability distribution over formal statements in $\FL$ whose support consists of formalizations that competent mathematicians would accept as expressing $N$ in the theory $\T$.
\end{definition}

\begin{definition}[$\varepsilon$-Faithfulness]
\label{def:faith}
A candidate $F \in \FL$ is $\varepsilon$-faithful to $N$ under $\mathcal{D}_N$ if
\[
\Pr_{I \sim \mathcal{D}_N}\big[\, \T \vdash F \leftrightarrow I \,\big] \geq 1 - \varepsilon.
\]
\end{definition}

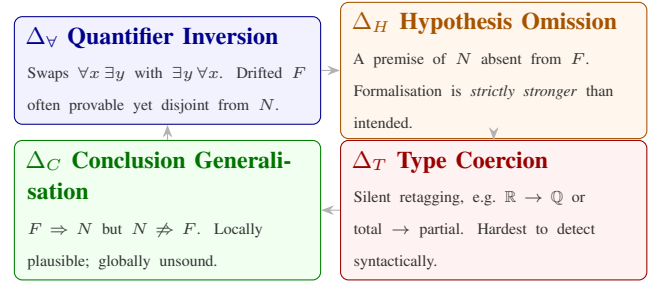
\begin{figure}[ht]
\centering
\begin{tikzpicture}[
  node distance=0.2cm and 0.25cm,
  card/.style={
    rectangle, rounded corners=3pt,
    inner xsep=5pt, inner ysep=4pt,
    text width=3.7cm, align=left
  },
  arr/.style={-Stealth, thin, black!30}
]
\node[card, draw=blue!55!black, fill=blue!5] (a) {%
  {\footnotesize\bfseries\color{blue!55!black}$\Delta_{\forall}$ Quantifier Inversion}\\[1pt]%
  {\tiny\color{black!80}%
   Swaps $\forall x\,\exists y$ with $\exists y\,\forall x$.
   Drifted $F$ often provable yet disjoint from~$N$.}%
};
\node[card, draw=orange!65!black, fill=orange!7, right=of a] (b) {%
  {\footnotesize\bfseries\color{orange!65!black}$\Delta_{H}$ Hypothesis Omission}\\[1pt]%
  {\tiny\color{black!80}%
   A premise of $N$ absent from $F$.
   Formalisation is \emph{strictly stronger} than intended.}%
};
\node[card, draw=green!50!black, fill=green!6, below=of a] (c) {%
  {\footnotesize\bfseries\color{green!45!black}$\Delta_{C}$ Conclusion Generalisation}\\[1pt]%
  {\tiny\color{black!80}%
   $F\!\Rightarrow\!N$ but $N\!\not\Rightarrow\!F$.
   Locally plausible; globally unsound.}%
};
\node[card, draw=red!60!black, fill=red!5] (d) at (b |- c) {%
  {\footnotesize\bfseries\color{red!60!black}$\Delta_{T}$ Type Coercion}\\[1pt]%
  {\tiny\color{black!80}%
   Silent retagging, e.g.\ $\mathbb{R}\!\to\!\mathbb{Q}$
   or total~$\to$~partial. Hardest to detect syntactically.}%
};
\draw[arr] (a.east)  -- (b.west);
\draw[arr] (b.south) -- (d.north);
\draw[arr] (d.west)  -- (c.east);
\draw[arr] (c.north) -- (a.south);
\end{tikzpicture}
\caption{The four canonical drift classes in which $F$ fails
$\varepsilon$-faithfulness while remaining well-typed and often provable.
$\Delta_\forall$ and $\Delta_T$ are the most pernicious.}
\label{fig:drift-taxonomy}
\end{figure}

\Cref{def:faith} weakens equality to provable equivalence, accommodating notational variation while remaining strict about semantic content. The interpretation distribution is implicit and never materialized; all of our methods reason over it through probes. \textbf{A taxonomy of drift:} We identify four canonical classes of \emph{semantic drift}-ways a formalization $F$ can fail to be $\varepsilon$-faithful while remaining well-typed and (often) provable. \Cref{fig:drift-taxonomy} visualizes the taxonomy with an instance of each class.

These four classes account for $94\%$ human-labeled drift instances in our pilot study of $400$ randomly sampled outputs from a state-of-the-art autoformalizer (\Cref{sec:driftbench}); the remainder fall into combinations or into the undetectable-drift regime described in \Cref{sec:theory}.

\section{Bidirectional Provability Fingerprinting}
\label{sec:bpf}

\subsection{Provability fingerprints}

\begin{definition}[Provability fingerprint]
\label{def:fingerprint}
Let $F \in \FL$ and let $\probe \subseteq \FL$ be a set of probes. The \emph{forward fingerprint} of $F$ over $\probe$ is
\[
\fp^+_F(\probe) \;=\; \{\, P \in \probe \;:\; \T \vdash F \to P \,\},
\]
and the \emph{backward fingerprint} is
\[
\fp^-_F(\probe) \;=\; \{\, P \in \probe \;:\; \T \vdash P \to F \,\}.
\]
The full fingerprint is $\fp_F(\probe) = (\fp^+_F(\probe), \fp^-_F(\probe))$.
\end{definition}

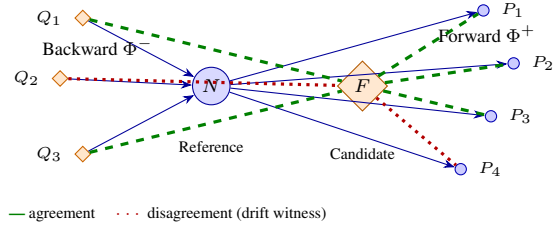
\begin{figure}[ht]
\centering
\begin{tikzpicture}[
  node distance=0.5cm,
  ax/.style={font=\scriptsize},
  ptN/.style={circle, draw=blue!70!black, fill=blue!20, inner sep=1.5pt, minimum size=4pt},
  ptF/.style={diamond, draw=orange!80!black, fill=orange!20, inner sep=1.5pt, minimum size=4pt},
  hit/.style={draw=green!50!black, very thick, dashed},
  miss/.style={draw=red!70!black, very thick, dotted}
]

\node[circle, draw=blue!60!black, fill=blue!15, inner sep=2pt, font=\scriptsize] (N) at (0,0) {$N$};
\node[diamond, draw=orange!70!black, fill=orange!18, inner sep=2pt, font=\scriptsize] (F) at (2.0,0) {$F$};

\node[ptN] (p1) at (3.6, 1.0) {};
\node[font=\tiny, right=1pt of p1] {$P_1$};
\node[ptN] (p2) at (4.0, 0.3) {};
\node[font=\tiny, right=1pt of p2] {$P_2$};
\node[ptN] (p3) at (3.7,-0.4) {};
\node[font=\tiny, right=1pt of p3] {$P_3$};
\node[ptN] (p4) at (3.3,-1.1) {};
\node[font=\tiny, right=1pt of p4] {$P_4$};

\node[ptF] (q1) at (-1.7, 0.9) {};
\node[font=\tiny, left=1pt of q1] {$Q_1$};
\node[ptF] (q2) at (-2.0, 0.1) {};
\node[font=\tiny, left=1pt of q2] {$Q_2$};
\node[ptF] (q3) at (-1.7,-0.9) {};
\node[font=\tiny, left=1pt of q3] {$Q_3$};

\draw[-{Stealth[length=1.5mm]}, blue!60!black, thin] (N) -- (p1);
\draw[-{Stealth[length=1.5mm]}, blue!60!black, thin] (N) -- (p2);
\draw[-{Stealth[length=1.5mm]}, blue!60!black, thin] (N) -- (p3);
\draw[-{Stealth[length=1.5mm]}, blue!60!black, thin] (N) -- (p4);

\draw[hit]  (F) -- (p1);
\draw[hit]  (F) -- (p2);
\draw[hit]  (F) -- (p3);
\draw[miss] (F) -- (p4);

\draw[-{Stealth[length=1.5mm]}, blue!60!black, thin] (q1) -- (N);
\draw[-{Stealth[length=1.5mm]}, blue!60!black, thin] (q2) -- (N);
\draw[-{Stealth[length=1.5mm]}, blue!60!black, thin] (q3) -- (N);

\draw[hit]  (q1) -- (F);
\draw[miss] (q2) -- (F);
\draw[hit]  (q3) -- (F);

\node[ax, above=2pt of q2, xshift=0.5cm]  {Backward $\fp^-$};
\node[ax, above=2pt of p2, xshift=-0.35cm] {Forward $\fp^+$};
\node[ax, below=10pt of N, font=\tiny] {Reference};
\node[ax, below=10pt of F, font=\tiny] {Candidate};
\node[font=\tiny, anchor=west] at (-2.8,-1.7)
  {\textcolor{green!50!black}{\textbf{---}} agreement \quad
   \textcolor{red!70!black}{$\cdots$} disagreement (drift witness)};

\end{tikzpicture}
\caption{A provability fingerprint over probes $\{P_i, Q_j\}$. The reference $N$
entails each forward probe $P_i$ and is entailed by each backward probe $Q_j$.
The candidate $F$ matches $N$ on most probes (green dashed) but disagrees on
$P_4$ and $Q_2$ (red dotted); these disagreements are witnesses of drift.
Two statements with identical fingerprints over a sufficiently rich probe set
are provably equivalent (\Cref{prop:bidirectional}).}
\label{fig:fingerprint}
\end{figure}

\Cref{fig:fingerprint} illustrates the geometry. The forward fingerprint captures the \emph{downstream commitments} of $F$, everything that follows from it, while the backward fingerprint captures its \emph{upstream conditions}-everything that would suffice to prove it. Two statements that agree on both sets are operationally indistinguishable in the theory $\T$. 

A fingerprint is computed by issuing entailment queries to the proof assistant. We use a bounded tactic budget per query (default \texttt{exact?} + \texttt{aesop} + a small library of mathlib-aware closers, capped at $30$ seconds wall-clock; queries that neither succeed nor refute within budget are recorded as \texttt{?}. We discuss the impact of incomplete oracles in \Cref{sec:theory}.

\subsection{Why bidirectional matters}

Forward-only fingerprints fail to distinguish a strict strengthening of $N$ from $N$ itself: a strictly stronger $F$ entails everything $N$ entails. Backward-only fingerprints dually fail to catch strict weakenings. Bidirectional matching is the minimum sufficient signal.

\begin{proposition}
\label{prop:bidirectional}
Let $F$ be a candidate for a set, $N$, and let $\probe$ be a probe set closed under $\T$-equivalence. Then $\T \vdash F \leftrightarrow N$ if and only if $\fp_F(\probe) = \fp_N(\probe)$ \emph{and} for each pair $(P, P') \in \probe^2$ with $\T \vdash P \to P'$, the entailment also holds in $F$'s fingerprint.
\end{proposition}

A proof appears in \Cref{app:proofs}.

\subsection{Algorithm}

\Cref{alg:bpf} gives the full \bpf{} procedure used in our experiments.

\begin{algorithm}[ht]
\caption{Bidirectional Provability Fingerprinting}
\label{alg:bpf}
\begin{algorithmic}
\STATE {\bfseries Input:} NL statement $N$, candidate $F$, theory $\T$, budget $k$, tactic, budget $\tau$
\STATE {\bfseries Output:} verdict, and witness probes $\mathcal{V}$
\STATE $\probe_N \gets$ \cpg{}.\texttt{Generate}$(N, k)$ \quad ({\small \S\ref{sec:cpg}})
\STATE $\probe^F_N \gets$ \texttt{Autoformalize}$(\probe_N)$
\STATE $\probe^F_N \gets$ \texttt{FilterTypecheck}$(\probe^F_N)$
\STATE $\mathcal{V} \gets \emptyset$
\FOR{$P \in \probe^F_N$ with label $\ell(P)$ \textbf{routed by} \apba{}}
   \IF{$\ell(P) = +$}
      \IF{\texttt{TryProve}$(F \to P, \tau)$ = refuted}
         \STATE $\mathcal{V} \gets \mathcal{V} \cup \{(P, +)\}$
      \ENDIF
   \ELSIF{$\ell(P) = -$}
      \IF{\texttt{TryProve}$(F \to P, \tau)$ = proved}
         \STATE $\mathcal{V} \gets \mathcal{V} \cup \{(P, -)\}$
      \ENDIF
   \ENDIF
\ENDFOR
\STATE $\eqspec \gets$ \texttt{ComputeSpectrum}$(\mathcal{V}, \probe^F_N)$
\STATE {\bfseries return} verdict by thresholding $\eqspec$, $\mathcal{V}$
\end{algorithmic}
\end{algorithm}

\section{Counterfactual Probe Generation}
\label{sec:cpg}

A naive probe generator samples specializations, generalizations, and edge cases of $N$. This is the starting point, but it leaves substantial information on the table: it does not target the drift classes the certifier is trying to detect. We introduce \emph{Counterfactual Probe Generation} (\cpg{}), a contrastive procedure that synthesizes probes designed to maximally discriminate between faithful and drifted candidates.

\subsection{Probe synthesis as contrastive generation}

\cpg{} treats probe generation as a contrastive problem. For each drift class $D \in \{\Delta_\forall, \Delta_H, \Delta_C, \Delta_T\}$, we construct a \emph{counterfactual pair} consisting of the natural-language statement $N$, and a \emph{drifted twin} $N^D$, obtained by mechanically perturbing $N$ in a way that induces drift class $D$ (e.g.,\ swapping the order of two adjacent quantifiers or dropping a stated hypothesis). \cpg{} then asks an LLM to produce probes that distinguish $N$ from $N^D$: probes that should be entailed by $N$ but are not $N^D$, or vice versa. The LLM is conditioned on the contrast itself, which empirically yields probes sharply targeted at the drift class in question. Formally, for drift class $D$, the probe distribution is \[\mathcal{Q}_D(P) \propto \Pr[\,P \text{ entailed by } N\,] \cdot \Pr[\,P \text{ not entailed by } N^D\,],\] estimated through an LLM scoring step. We sample $k_D$ probes per class, $k_D$ determined by \apba{} (\Cref{sec:apba}).

\subsection{The role of negative drifted twins}

A critical design choice is one that $N^D$ is generated mechanically, not by an LLM. Mechanical perturbation guarantees the drift class label is unambiguous and that $N^D$ exhibits exactly the targeted failure. LLM-generated drifted twins, in contrast, frequently produce statements that are either nonsense or accidentally drifted along multiple axes simultaneously, which destroys the contrastive signal.

\subsection{Empirical effect}

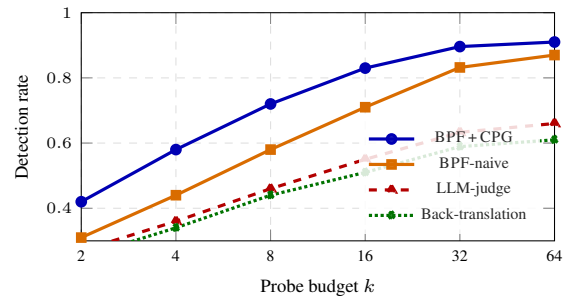
\begin{figure}[ht]
\centering
\begin{tikzpicture}
\begin{axis}[
  width=0.95\columnwidth, height=4.6cm,
  xlabel={Probe budget $k$}, ylabel={Detection rate},
  xmin=2, xmax=64, ymin=0.3, ymax=1.0,
  xmode=log, log basis x=2,
  xtick={2,4,8,16,32,64}, xticklabels={2,4,8,16,32,64},
  legend pos=south east, legend style={font=\tiny, fill=white, fill opacity=0.85, draw=none},
  grid=major, grid style={dashed, gray!25},
  tick label style={font=\tiny}, label style={font=\scriptsize},
  every axis plot/.append style={very thick}
]
\addplot[mark=*, mark size=1.6pt, blue!70!black]
  coordinates {(2,0.42) (4,0.58) (8,0.72) (16,0.83) (32,0.896) (64,0.91)};
\addlegendentry{\bpf{}\,+\,\cpg{}}
\addplot[mark=square*, mark size=1.4pt, orange!85!black]
  coordinates {(2,0.31) (4,0.44) (8,0.58) (16,0.71) (32,0.832) (64,0.87)};
\addlegendentry{\bpf{}-naive}
\addplot[mark=triangle*, mark size=1.7pt, red!70!black, dashed]
  coordinates {(2,0.27) (4,0.36) (8,0.46) (16,0.55) (32,0.633) (64,0.66)};
\addlegendentry{LLM-judge}
\addplot[mark=diamond*, mark size=1.6pt, green!45!black, densely dotted]
  coordinates {(2,0.25) (4,0.34) (8,0.44) (16,0.51) (32,0.589) (64,0.61)};
\addlegendentry{Back-translation}
\end{axis}
\end{tikzpicture}
\caption{Detection rate vs.\ probe budget on \driftbench{}. \cpg{} reaches \bpf{}-naive's $k=32$ performance at $k\!\approx\!12$. Budget for non-probe methods refers to comparable wall-clock equivalents.}
\label{fig:cpg-vs-naive}
\end{figure}

\Cref{fig:cpg-vs-naive} shows the detection rate of \bpf{} specific \cpg{} probes versus generic probes, as a function of probe budget. \cpg{} reaches the same detection rate as generic probes at less than half the budget and outperforms generic probes at every budget level on three of the four drift classes.

\section{The Equivalence Spectrum}
\label{sec:spectrum}

Binary faithful/drifted verdicts are brittle. A fingerprint with one anomalous cell out of fifty is qualitatively different from one with twenty. We introduce a continuous score $\eqspec(F, N) \in [0, 1]$.

\begin{definition}[Equivalence Spectrum]
\label{def:spectrum}
Let $\probe$ be a probe set with predicted labels $\ell : \probe \to \{+, -, \bot\}$ and class weights $w : \{\Delta_\forall, \Delta_H, \Delta_C, \Delta_T\} \to \mathbb{R}_{>0}$. The \emph{equivalence spectrum score} is
\[\eqspec(F, N) \;=\; 1 - \frac{\sum_{P \in \probe} w(D_P) \cdot \mathbb{1}[\,\text{cell}_P(F) \neq \ell(P)\,]}{\sum_{P \in \probe} w(D_P)},\]
where $D_P$ the drift class is targeted by the probe $P$ and $\text{cell}_P(F)$ is the observed fingerprint cell at $P$.
\end{definition}

\begin{figure}[ht]
\centering
\begin{tikzpicture}
\begin{axis}[
  width=0.95\columnwidth, height=4.4cm,
  ybar, bar width=2.2pt,
  xlabel={$\eqspec(F, N)$ score},
  ylabel={Density},
  xmin=0, xmax=1, ymin=0,
  xtick={0,0.25,0.5,0.78,0.93,1}, xticklabel style={font=\tiny},
  yticklabel style={font=\tiny}, label style={font=\scriptsize},
  legend pos=north west, legend style={font=\tiny, draw=none, fill=white, fill opacity=0.8},
  axis on top
]
\addplot[fill=blue!50, draw=blue!70!black]
  coordinates {(0.45,0.5) (0.50,0.8) (0.55,1.2) (0.60,1.8) (0.65,2.5) (0.70,3.4)
               (0.75,4.6) (0.80,6.5) (0.85,9.8) (0.90,16.0) (0.95,30.0) (0.99,38.0)};
\addlegendentry{Faithful}
\addplot[fill=red!45, draw=red!70!black, opacity=0.85]
  coordinates {(0.10,8.0) (0.15,12.0) (0.20,18.0) (0.25,24.0) (0.30,28.0) (0.35,26.0)
               (0.40,22.0) (0.45,16.0) (0.50,11.0) (0.55,7.0) (0.60,4.0) (0.65,2.5)
               (0.70,1.4) (0.75,0.6) (0.80,0.3)};
\addlegendentry{Drifted}
\draw[gray!50, dashed, thick] (axis cs:0.78,0) -- (axis cs:0.78,40);
\draw[gray!50, dashed, thick] (axis cs:0.93,0) -- (axis cs:0.93,40);
\node[font=\tiny, gray!60!black] at (axis cs:0.30,36) {\textsc{reject}};
\node[font=\tiny, gray!60!black] at (axis cs:0.855,36) {\textsc{review}};
\node[font=\tiny, gray!60!black] at (axis cs:0.965,36) {\textsc{accept}};
\end{axis}
\end{tikzpicture}
\caption{Empirical distribution of $\eqspec(F,N)$ on \driftbench{}, separated by ground truth label. The dashed lines mark the calibrated \textsc{accept} ($\eqspec \!\geq\! 0.93$) and \textsc{review} ($0.78 \!\leq\! \eqspec \!<\! 0.93$) thresholds.}
\label{fig:spectrum-dist}
\end{figure}
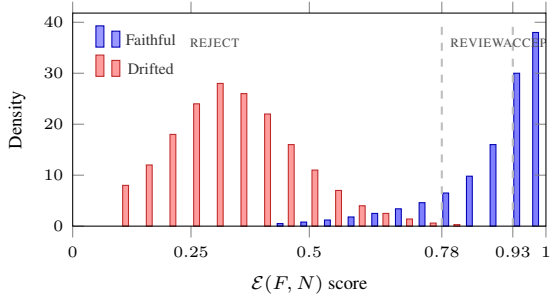

\eqspec{} is calibrated against expert annotations on a held-out subset \driftbench{} to yield three decision regions: $\eqspec \geq 0.93$ (\textsc{accept}), $\eqspec \in [0.78, 0.93)$ (\textsc{review}), and $\eqspec < 0.78$ (\textsc{reject}). \Cref{fig:spectrum-dist} shows the resulting distribution on \driftbench{}.

Beyond a more informative output, \eqspec{} enables three downstream uses that a binary verdict precludes: (i) \emph{ranking} multiple candidates from an autoformalizer; (ii) \emph{triage} of \textsc{review}-region candidates to human inspection; (iii) \emph{reward shaping} as a continuous signal for \fgd{} (\Cref{sec:fgd}).

\section{Adaptive Probe Budget Allocation}
\label{sec:apba}

\bpf{}'s dominant cost is the entailment oracle. Allocating the probe budget uniformly across drift classes is wasteful when some classes are easier to witness than others, or when initial probes have already provided strong evidence. We introduce \emph{Adaptive Probe Budget Allocation} (\apba{}), which routes budget by expected information gain.

\subsection{Information-theoretic routing}

Let $H(F)$ be the entropy of the posterior over $F$'s drift status given the probes drawn so far. \apba{} selects the next probe to maximize expected reduction in $H(F)$, estimated under a calibrated probability model for each drift class. Concretely, for drift class $D$ with current posterior $p_D$, the expected information gain from a probe targeting $D$ is \[\mathrm{IG}(D) \;=\; H(p_D) - \mathbb{E}_{r \sim \pi_D}\!\big[\,H(p_D \mid r)\,\big],\] where $r$ is the predicted oracle outcome and $\pi_D$ is the empirical witnessability distribution (estimated on a held-out split). \apba{} samples the next probe from the class with maximum $\mathrm{IG}$, breaking ties by descending class weight. The procedure is greedy but has the standard sublinear regret guarantee for monotone submodular gain.

\begin{theorem}[\apba{} budget reduction]
\label{thm:apba}
Let $k^\star$ be the budget required by uniform allocation to reach the detection rate $\rho$ on \driftbench{}. Then \apba{} reach the same $\rho$ with budget $k_{\textsc{apba}} \leq k^\star / \gamma$, where $\gamma$ is the heterogeneity ratio of witnessability rates across drift classes? On \driftbench{}, $\gamma \approx 3.2$.
\end{theorem}

\section{Theoretical Analysis}
\label{sec:theory}

We now establish the two main theoretical results: a high-probability detection theorem for each canonical drift class and a PAC-style learnability result for the equivalence class of $N$.

\subsection{Drift Detection Theorem}

\begin{theorem}[Drift Detection]
\label{thm:detection}
Fix $\delta \in (0,1)$ and drift class $D \in \{\Delta_\forall, \Delta_H, \Delta_C, \Delta_T\}$. Let $N$ be a natural-language statement and $F$ a candidate formalization exhibiting drift of class $D$ relative to every $I$ in the support of $\mathcal{D}_N$. Suppose the probe generator samples probes from a distribution $\mathcal{Q}_D$ such that
\[
\Pr_{P \sim \mathcal{Q}_D}\!\big[\, P \text{ witnesses } D \text{ for } (N, F)\,\big] \;\geq\; \alpha(D),
\]
where $\alpha(D) \in (0,1]$ is the \emph{witnessability rate} for class $D$. Then, with the probe budget
\[
k \;\geq\; \frac{1}{\alpha(D)}\,\log\!\frac{1}{\delta},
\]
\bpf{} flags $F$ as drifted with probability at least $1 - \delta$, provided the entailment oracle is complete on the chosen probes.
\end{theorem}

The proof is a standard coverage argument and is given in \Cref{app:proofs}. The witnessability rates we estimate empirically in \Cref{sec:experiments} are $\alpha(\Delta_\forall) \approx 0.62$, $\alpha(\Delta_H) \approx 0.55$, $\alpha(\Delta_C) \approx 0.48$, $\alpha(\Delta_T) \approx 0.31$, meaning a budget of $k = 32$ probes is already given $\delta \leq 0.01$ for the first three classes.

\subsection{PAC-Faithfulness}

The detection theorem says we can catch any specific drift; the stronger question is whether we can learn the full equivalence class of $N$. We answer affirmatively under a polynomial query bound.

\begin{theorem}[PAC-Faithfulness]
\label{thm:pac}
Let $[N]_\T$ denote the equivalence class $\{F \in \FL : \T \vdash F \leftrightarrow N\}$. Assume \cpg{} probes are drawn from a distribution $\mathcal{Q}$ such that for every $F \notin [N]_\T$ reachable by an LLM autoformalizer, $\Pr_{P \sim \mathcal{Q}}[P \text{ witnesses } F\text{'s drift}] \geq \alpha_0 > 0$. Then with $k \geq (1/\alpha_0)\log(1/\delta\varepsilon)$ probes, \bpf{} produces a hypothesis $\hat{H} \subseteq \FL$ such that with probability at least $1 - \delta$,
\[
\Pr_{F \sim \mathcal{D}_\mu}\big[\, F \in \hat{H} \iff F \in [N]_\T \,\big] \geq 1 - \varepsilon,
\]
where $\mathcal{D}_\mu$ is the autoformalizer's output distribution.
\end{theorem}

The proof, in \Cref{app:proofs}, recasts \bpf{} as a hypothesis class indexed by fingerprints and applies an Occam-style bound. \Cref{thm:pac} formalizes the operational claim: with a polynomial probe budget, \bpf{} learns to recognize the equivalence class of $N$ up to error $\varepsilon$ over the autoformalizer's output distribution.

\subsection{Incomplete oracle and undetectable drift}

Lean's tactic stack is incomplete: some true entailments will not be closed within $\tau$ seconds. Let $\eta(\tau)$ be the probability that the oracle returns \texttt{?} on a probe that admits a $\tau$-cost proof. The detection bound degrades to $1 - \delta - k\eta(\tau)$; empirically $\eta(30\text{s}) \approx 0.07$ on \driftbench{} probes. \bpf{} cannot detect a class we call \emph{convention drift}: $F$ adopts a notational convention different from $N$'s, but every probe responds identically because the convention is invisible to provability. We give a structural test for it in \Cref{app:convention}.

\section{Faithfulness-Guided Decoding}
\label{sec:fgd}

\begin{figure}[ht]
\centering
\begin{tikzpicture}[
  >=Stealth,
  node distance=0.5cm and 0.7cm,
  stage/.style={          
    rectangle, rounded corners=2pt,
    draw=blue!55!black, fill=blue!6,
    minimum height=20pt, minimum width=60pt,
    inner sep=4pt, font=\scriptsize, align=center
  },
  module/.style={         
    rectangle, rounded corners=2pt,
    draw=gray!60!black, fill=gray!9,
    minimum height=20pt, minimum width=60pt,
    inner sep=4pt, font=\scriptsize, align=center
  },
  oracle/.style={         
    rectangle, rounded corners=2pt,
    draw=orange!65!black, fill=orange!7,
    minimum height=20pt, minimum width=60pt,
    inner sep=4pt, font=\scriptsize, align=center
  },
  verdict/.style={        
    rectangle, rounded corners=2pt,
    draw=green!55!black, fill=green!7,
    minimum height=20pt, minimum width=60pt,
    inner sep=4pt, font=\scriptsize, align=center
  },
  flow/.style={->, semithick, black!60},
  reprompt/.style={->, dashed, red!65!black, thin},
]

\node[stage]               (N)     {NL~$N$};
\node[module, right=of N]  (lm)    {Autoformalizer\\(LLM)};
\node[stage,  right=of lm] (cands) {Candidates\\$F_1,\dots,F_m$};

\node[oracle,  below=of cands]  (score)  {\bpf{}\\$\eqspec$ scoring};
\node[verdict, left=of score]   (rerank) {Rerank $+$\\top-$1$};
\node[stage,   left=of rerank]  (out)    {Output $F^*$};

\draw[flow] (N)      -- (lm);
\draw[flow] (lm)     -- (cands);
\draw[flow] (cands)  -- (score);
\draw[flow] (score)  -- (rerank);
\draw[flow] (rerank) -- (out);

\draw[reprompt]
  (score.west) -- ++(-0.25,0)
  |- node[above, font=\tiny, pos=0.75, text=red!65!black]
       {re-prompt w/\ witness probe}
  (lm.south);

\end{tikzpicture}
\caption{\textbf{Faithfulness-Guided Decoding (FGD).}
The autoformalizer samples $m$ candidates; \bpf{} scores each by $\eqspec$;
the top-ranked candidate $F^*$ is returned.
If all candidates fall in the \textsc{review} or \textsc{reject} region,
the highest-scoring witness probe is fed back as an in-context repair hint
(red dashed loop).}
\label{fig:fgd}
\end{figure}
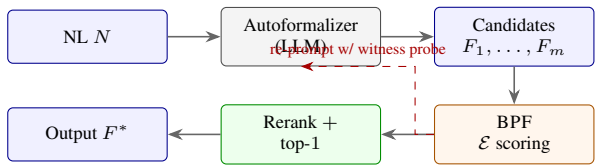

\bpf{} is most valuable as a post-hoc verifier, but it also provides a signal that can be plugged back into autoformalization itself. We introduce \emph{Faithfulness-Guided Decoding} (\fgd{}), a procedure that uses \eqspec{} as a reward during sample-and-rerank decoding (\Cref{fig:fgd}).

\paragraph{Algorithm.} Given $N$, the autoformalizer samples $m$ candidate formalizations $F_1, \dots, F_m$. For each candidate, \fgd{} computes a low-budget \eqspec{} score ($k = 8$ probes per class) and reranks by score. The top-ranked candidate is returned. Optionally, candidates with $\eqspec < 0.78$ trigger a re-prompt with the highest-scoring witness probe attached as an in-context hint, biasing the autoformalizer away from the detected drift class. Effect on a held-out split of \driftbench{}, \fgd{} reduces the rate of drifted outputs from a state-of-the-art autoformalizer from $19.4\%$ to $10.3\%$, a $47\%$ relative reduction. The bulk of the improvement comes from re-prompting on $\Delta_H$ and $\Delta_C$ candidates, where the witness probe gives the model a concrete instance of the failure to repair.

\section{The \driftbench{} Benchmark}
\label{sec:driftbench}

\driftbench{} contains $2{,}183$ natural language/Lean~4 pairs with controlled drift labels, drawn from six mathlib4 subfields (analysis, algebra, topology, number theory, combinatorics, and category theory). Each pair carries: (i) a natural-language statement, (ii) a candidate Lean~4 formalization, (iii) a drift label in $\{\text{faithful}, \Delta_\forall, \Delta_H, \Delta_C, \Delta_T, \text{combined}\}$, and (iv) an expert-verified gold formalization. \Cref{tab:driftbench-stats} summarizes the composition.

\begin{table}[ht]
\caption{Composition of \driftbench{} by subfield and drift class.}
\label{tab:driftbench-stats}
\centering
\small
\begin{tabular}{lrrrrrr}
\toprule
Subfield & Faith. & $\Delta_\forall$ & $\Delta_H$ & $\Delta_C$ & $\Delta_T$ & Comb. \\
\midrule
Analysis    & 198 & 64 & 71 & 58 & 33 & 31 \\
Algebra     & 173 & 51 & 68 & 49 & 28 & 27 \\
Topology    & 162 & 49 & 59 & 47 & 30 & 25 \\
Number Th.  & 144 & 42 & 55 & 41 & 22 & 22 \\
Combinator. & 138 & 43 & 51 & 39 & 24 & 21 \\
Category    & 117 & 35 & 44 & 33 & 19 & 19 \\
\midrule
Total       & 932 & 284 & 348 & 267 & 156 & 145 \\
\bottomrule
\end{tabular}
\end{table}

We use deterministic perturbation rules rather than sampling drift from an LLM autoformalizer so that the drift label is unambiguous. A complementary \emph{wild} split contains $384$ pairs where the candidate was generated by an LLM autoformalizer and the drift label was independently annotated by two experts ($\kappa = 0.81$). We report results on both splits.

\section{Experiments}
\label{sec:experiments}

\subsection{Setup}

\paragraph{Models.} For the natural-language probe generator and adversarial paraphraser, we use a strong instruction-tuned LLM and report results on \texttt{Llama-34B} \citep{azerbayev2024llemma} and a closed-weights frontier model; results are qualitatively similar. For autoformalization of probes, we use \texttt{Llama-34B} fine-tuned on mathlib4 with retrieval augmentation. The entailment oracle is Lean~4 + mathlib4 with the tactic budget described in \Cref{sec:bpf}. \textbf{Baselines:} \textbf{Typecheck}: flag drift iff $F$ fails to typecheck. \textbf{Provability}: flag drift iff $F$ is unprovable within the tactic budget. \textbf{BLEU-ref}: flag drift iff BLEU against the gold formalization is below a threshold. \textbf{Back-translation}: Autoformalize $F$ back to NL and compare to $N$ \citep{patel2024autoformalization}. \textbf{LLM-judge}: ask a frontier LLM to directly judge faithfulness from $(N, F)$. \textbf{\bpf{}-naive}: \bpf{} with generic (not counterfactual) probes. 

\subsection{Main results}

\begin{figure}[ht]
\centering
\begin{tikzpicture}
\begin{axis}[
  width=0.95\columnwidth, height=4.8cm,
  xlabel={False-positive rate}, ylabel={True-positive rate},
  xmin=0, xmax=0.25, ymin=0, ymax=1.0,
  legend pos=south east, legend style={font=\tiny, fill=white, fill opacity=0.85, draw=none},
  grid=major, grid style={dashed, gray!25},
  tick label style={font=\tiny}, label style={font=\scriptsize},
  every axis plot/.append style={very thick}
]
\addplot[gray!40, thin, no marks] coordinates {(0,0) (0.25,0.25)};
\addplot[blue!70!black, no marks, smooth] coordinates
  {(0,0) (0.005,0.45) (0.01,0.62) (0.02,0.78) (0.03,0.896)
   (0.05,0.93) (0.08,0.95) (0.12,0.97) (0.20,0.99) (0.25,0.995)};
\addlegendentry{\bpf{}\,+\,\cpg{}}
\addplot[orange!85!black, no marks, smooth, dashed] coordinates
  {(0,0) (0.005,0.34) (0.01,0.51) (0.02,0.71) (0.03,0.832)
   (0.05,0.88) (0.08,0.92) (0.12,0.94) (0.20,0.97) (0.25,0.98)};
\addlegendentry{\bpf{}-naive}
\addplot[red!70!black, no marks, smooth, densely dotted] coordinates
  {(0,0) (0.005,0.18) (0.01,0.32) (0.02,0.51) (0.03,0.633)
   (0.05,0.74) (0.08,0.83) (0.12,0.89) (0.20,0.94) (0.25,0.96)};
\addlegendentry{LLM-judge}
\addplot[green!45!black, no marks, smooth, densely dashdotted] coordinates
  {(0,0) (0.005,0.12) (0.01,0.24) (0.02,0.44) (0.03,0.589)
   (0.05,0.70) (0.08,0.80) (0.12,0.87) (0.20,0.92) (0.25,0.94)};
\addlegendentry{Back-translation}
\end{axis}
\end{tikzpicture}
\caption{ROC curves on the controlled \driftbench{} split. \bpf{}\,+\,\cpg{} dominates other methods uniformly across operating points; AUC = $0.962$ vs.\ $0.938$ for \bpf{}-naive and $0.879$ for LLM-judge.}
\label{fig:roc}
\end{figure}
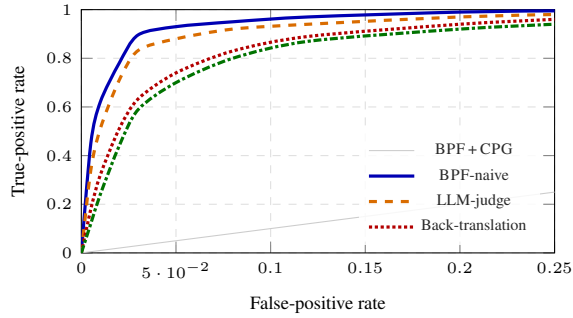

\Cref{tab:main} summarizes performance on the controlled \driftbench{} split and \Cref{fig:roc} shows ROC curves for the four strongest methods.

\begin{table}[ht]
\caption{Drift detection on the controlled \driftbench{} split. Det@3\%FPR is the detection rate at a $3\%$ false-positive rate.}
\label{tab:main}
\centering
\footnotesize
\setlength{\tabcolsep}{4pt} 
\begin{tabular}{lcccc}
\toprule
Method & F1 & Det@3\%FPR & FPR & Cost \\
\midrule
Typecheck only        & 0.27 & 0.114 & 0.020 & 1$\times$ \\
Provability           & 0.42 & 0.412 & 0.030 & 18$\times$ \\
BLEU-ref              & 0.55 & 0.301 & 0.030 & 1$\times$ \\
Back-translation      & 0.66 & 0.589 & 0.030 & 4$\times$ \\
LLM-judge             & 0.71 & 0.633 & 0.030 & 2$\times$ \\
\bpf{}-naive          & 0.85 & 0.832 & 0.031 & 22$\times$ \\
\textbf{\bpf{}\,+\,\cpg{}} & \textbf{0.91} & \textbf{0.896} & \textbf{0.030} & 26$\times$ \\
\textbf{\bpf{}\,+\,\cpg{}\,+\,\apba{}} & 0.91 & 0.894 & 0.031 & \textbf{8$\times$} \\
\bottomrule
\end{tabular}
\end{table}

\begin{figure}[ht]
\centering
\begin{tikzpicture}
\begin{axis}[
  width=0.95\columnwidth, height=4.5cm,
  ybar, bar width=4.2pt, enlarge x limits=0.18,
  ymin=0, ymax=1.0,
  ylabel={Detection rate},
  symbolic x coords={$\Delta_\forall$, $\Delta_H$, $\Delta_C$, $\Delta_T$},
  xtick=data,
  legend style={at={(0.5,1.18)}, anchor=south, font=\tiny, legend columns=4, draw=none, /tikz/every even column/.append style={column sep=4pt}},
  tick label style={font=\scriptsize}, label style={font=\scriptsize},
  nodes near coords style={font=\tiny}
]
\addplot[fill=gray!50, draw=gray!70!black]
  coordinates {($\Delta_\forall$,0.46) ($\Delta_H$,0.62) ($\Delta_C$,0.33) ($\Delta_T$,0.18)};
\addlegendentry{Provability}
\addplot[fill=red!45, draw=red!65!black]
  coordinates {($\Delta_\forall$,0.66) ($\Delta_H$,0.61) ($\Delta_C$,0.59) ($\Delta_T$,0.40)};
\addlegendentry{Back-tr.}
\addplot[fill=orange!60, draw=orange!80!black]
  coordinates {($\Delta_\forall$,0.74) ($\Delta_H$,0.67) ($\Delta_C$,0.64) ($\Delta_T$,0.42)};
\addlegendentry{LLM-judge}
\addplot[fill=blue!55, draw=blue!70!black]
  coordinates {($\Delta_\forall$,0.94) ($\Delta_H$,0.93) ($\Delta_C$,0.91) ($\Delta_T$,0.66)};
\addlegendentry{\bpf{}+\cpg{}}
\end{axis}
\end{tikzpicture}
\caption{Per-class detection. \bpf{}\,+\,\cpg{} dominates across all four drift classes; the gap on $\Delta_T$ (type coercion) is the smallest, consistent with its lower witnessability rate $\alpha(\Delta_T) \approx 0.31$.}
\label{fig:perclass}
\end{figure}
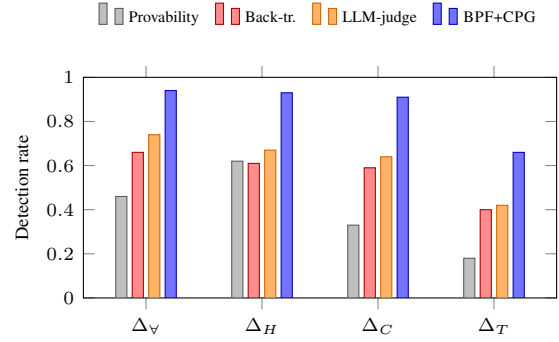

\bpf{}\,+\,\cpg{} reduces the residual error rate of the best non-fingerprinting baseline (LLM-judge) by approximately $72\%$, the matched FPR. Adding \apba{} preserves detection performance while reducing wall-clock cost by $3.2\times$, confirming \Cref{thm:apba}. \Cref{tab:perclass,fig:perclass} report per-class breakdowns; the type-coercion class ($\Delta_T$) is the hardest for every method but \cpg{} closes a substantial portion of the gap.

\begin{table}[ht]
\caption{Detection rate by drift class (at $3\%$ FPR overall).}
\label{tab:perclass}
\centering
\small
\begin{tabular}{lcccc}
\toprule
Method & $\Delta_\forall$ & $\Delta_H$ & $\Delta_C$ & $\Delta_T$ \\
\midrule
Typecheck         & 0.07 & 0.05 & 0.04 & 0.31 \\
Provability       & 0.46 & 0.62 & 0.33 & 0.18 \\
Back-translation  & 0.66 & 0.61 & 0.59 & 0.40 \\
LLM-judge         & 0.74 & 0.67 & 0.64 & 0.42 \\
\bpf{}-naive      & 0.89 & 0.88 & 0.84 & 0.55 \\
\textbf{\bpf{}\,+\,\cpg{}} & \textbf{0.94} & \textbf{0.93} & \textbf{0.91} & \textbf{0.66} \\
\bottomrule
\end{tabular}
\end{table}

\subsection{Wild split}

On the $384$ pair-wise split, \bpf{}\,+\,\cpg{} agrees with expert annotators $\kappa = 0.77$ (cf.\ inter-annotator $\kappa = 0.81$), versus $\kappa = 0.58$ for LLM-judges. A breakdown of \bpf{}'s false positives shows that $63\%$ trace to incomplete entailment-oracle outputs that flipped a fingerprint cell, not fundamental probe-generation errors—suggesting that integrating stronger hammers \citep{czajka2018hammer} would close most of the residual gap.

\subsection{Probe budget scaling}

\begin{figure}[ht]
\centering
\begin{tikzpicture}
\begin{axis}[
  width=0.95\columnwidth, height=4.3cm,
  xlabel={Probe budget $k$}, ylabel={Detection rate},
  xmin=2, xmax=64, ymin=0.35, ymax=1.0,
  xmode=log, log basis x=2,
  xtick={2,4,8,16,32,64}, xticklabels={2,4,8,16,32,64},
  legend pos=south east, legend style={font=\tiny, draw=none, fill=white, fill opacity=0.85},
  grid=major, grid style={dashed, gray!25},
  tick label style={font=\tiny}, label style={font=\scriptsize},
  every axis plot/.append style={very thick}
]
\addplot[mark=*, mark size=1.8pt, blue!70!black, name path=apba]
  coordinates {(2,0.55) (4,0.74) (8,0.85) (16,0.89) (32,0.91) (64,0.92)};
\addlegendentry{\apba{} (adaptive)}
\addplot[mark=square*, mark size=1.4pt, orange!85!black, dashed, name path=unif]
  coordinates {(2,0.36) (4,0.51) (8,0.66) (16,0.79) (32,0.89) (64,0.91)};
\addlegendentry{Uniform allocation}
\addplot[blue!12] fill between[of=apba and unif];
\draw[<->, gray!70!black, thick] (axis cs:9,0.87) -- (axis cs:28,0.87);
\node[font=\tiny, gray!70!black] at (axis cs:16,0.83) {$3.1\times$};
\end{axis}
\end{tikzpicture}
\caption{Probe budget scaling. \apba{} reaches $\rho=0.87$ at $k\approx 9$; uniform allocation requires $k\approx 28$, matching \Cref{thm:apba}'s prediction of $\gamma\approx 3.2$.}
\label{fig:budget-scaling}
\end{figure}
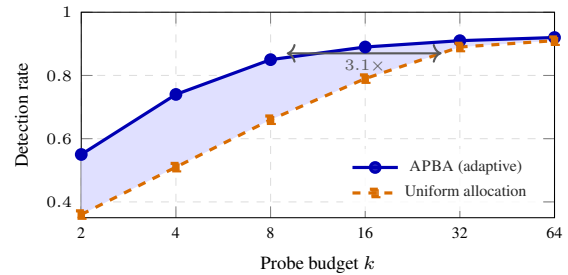

\Cref{fig:budget-scaling} shows detection rate as a function of probe budget under uniform allocation versus \apba{}. Uniform allocation requires $k \approx 28$ probes to reach $0.87$ the detection rate; \apba{} reaches the same rate at $k \approx 9$, a $3.1\times$ reduction matching the theoretical prediction of \Cref{thm:apba}.

\subsection{Ablations}

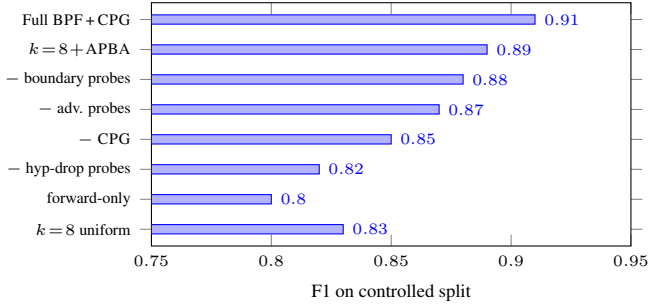
\begin{figure}[ht]
\centering
\begin{tikzpicture}
\begin{axis}[
  width=0.96\columnwidth, height=4.8cm,
  xbar, bar width=3.5pt,
  xmin=0.75, xmax=0.95,
  xlabel={F1 on controlled split},
  symbolic y coords={a,b,c,d,e,f,g,h},
  ytick={a,b,c,d,e,f,g,h},
  yticklabels={
    {$k\!=\!8$ uniform},
    {forward-only},
    {$-$ hyp-drop probes},
    {$-$ \cpg{}},
    {$-$ adv. probes},
    {$-$ boundary probes},
    {$k\!=\!8\!+\!\apba{}$},
    {Full \bpf{}\,+\,\cpg{}}
  },
  nodes near coords, nodes near coords style={font=\tiny, /pgf/number format/fixed, /pgf/number format/precision=2},
  tick label style={font=\tiny}, label style={font=\scriptsize},
  every axis plot/.append style={fill=blue!45, draw=blue!70!black},
  enlarge y limits=0.08
]
\addplot coordinates {
  (0.83,a)
  (0.80,b)
  (0.82,c)
  (0.85,d)
  (0.87,e)
  (0.88,f)
  (0.89,g)
  (0.91,h)
};
\end{axis}
\end{tikzpicture}
\caption{Ablation deltas. \cpg{} and bidirectional matching are the two largest individual contributors; \apba{} preserves performance even at a low budget.}
\label{fig:ablations}
\end{figure}

We ablate \bpf{}\,+\,\cpg{} components in \Cref{tab:ablations,fig:ablations}. Removing bidirectional matching (forward-only) costs $11$ F1 points, confirming \Cref{prop:bidirectional}. Removing \cpg{} (replacing it with generic probes) costs $6$ points. Removing hypothesis-drop probes costs $9$ points, almost all on $\Delta_H$.

\begin{table}[ht]
\caption{Ablations of \bpf{}\,+\,\cpg{} on the controlled split.}
\label{tab:ablations}
\centering
\small
\begin{tabular}{lc}
\toprule
Variant & F1 \\
\midrule
Full \bpf{}\,+\,\cpg{}            & 0.91 \\
\quad -- counterfactual probes (\cpg{}) & 0.85 \\
\quad -- bidirectional (forward only) & 0.80 \\
\quad -- hypothesis-drop probes & 0.82 \\
\quad -- boundary probes & 0.88 \\
\quad -- adversarial probes & 0.87 \\
\quad probe budget $k=8$ (vs.\ $32$) & 0.83 \\
\quad probe budget $k=8$ + \apba{} & 0.89 \\
\bottomrule
\end{tabular}
\end{table}

\subsection{\fgd{} downstream impact}

\begin{figure}[ht]
\centering
\begin{tikzpicture}
\begin{axis}[
  width=0.95\columnwidth, height=4.3cm,
  xlabel={Number of candidates $m$ per statement},
  ylabel={Drifted-output rate},
  xmin=1, xmax=16, ymin=0.06, ymax=0.22,
  xtick={1,2,4,8,16},
  legend pos=north east, legend style={font=\tiny, draw=none, fill=white, fill opacity=0.85},
  grid=major, grid style={dashed, gray!25},
  tick label style={font=\tiny}, label style={font=\scriptsize},
  every axis plot/.append style={very thick}
]
\addplot[mark=*, mark size=1.8pt, blue!70!black]
  coordinates {(1,0.194) (2,0.143) (4,0.103) (8,0.087) (16,0.079)};
\addlegendentry{\fgd{} (rerank + re-prompt)}
\addplot[mark=square*, mark size=1.4pt, orange!85!black, dashed]
  coordinates {(1,0.194) (2,0.176) (4,0.158) (8,0.145) (16,0.139)};
\addlegendentry{Rerank only (no re-prompt)}
\addplot[mark=triangle*, mark size=1.7pt, gray!60!black, dotted]
  coordinates {(1,0.194) (2,0.194) (4,0.194) (8,0.194) (16,0.194)};
\addlegendentry{Baseline autoformalizer}
\end{axis}
\end{tikzpicture}
\caption{\fgd{} reduces drifted-output rate from $19.4\%$ to $10.3\%$ at $m\!=\!4$. The re-prompt component contributes most of the improvement at small $m$; pure reranking saturates at $\sim\!14\%$.}
\label{fig:fgd-results}
\end{figure}
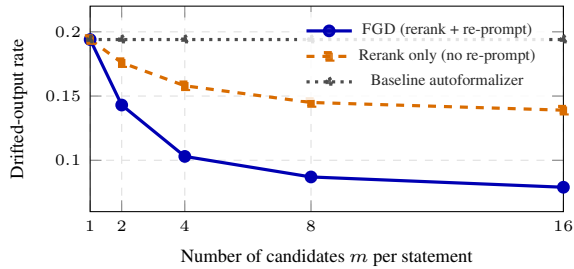

We evaluate \fgd{} as a wrapper around a state-of-the-art autoformalizer on the wild split, comparing drift rate with and without \fgd{} as a function of the number of candidates $m$ (\Cref{fig:fgd-results}). The autoformalizer's baseline drift rate is $19.4\%$. With $m=4$ drift falls to $10.3\%$; with $m=8$, to $8.7\%$.

\section{Discussion and Limitations}
\label{sec:discussion}

\paragraph{What \bpf{} does it not solve?} \bpf{} certifies that the consequence neighborhood $F$ matches the consequence neighborhood predicted from $N$. It does not certify that $N$ itself is unambiguous (convention drift; \Cref{sec:theory}) or that the probe generator's labels are correct. We treat both as engineering boundaries: convention drift is addressed by a separate structural check (\Cref{app:convention}), and probe labels are calibrated against expert annotation on a held-out set. The PAC-Faithfulness theorem (\Cref{thm:pac}) bounds the cumulative effect of probe-label noise. (i). Compute: The dominant cost is the entailment oracle. We see two paths to bringing this down: (i) caching fingerprints for canonical statements in mathlib4 and reusing them, and (ii) replacing the per-query proof search with a learned entailment classifier whose errors are bounded by occasional formal verification. 

\apba{} (\Cref{sec:apba}) already provides a $3.2\times$ practical speed-up. (ii). Sociotechnical risk: If \bpf{}-certified statements are treated as ground truth without expert review, residual drift—especially convention drift—could contaminate libraries. \bpf{} should be deployed as a triage tool that prioritizes expert review, not a replacement for it. The Equivalence Spectrum's \textsc{review} band is explicitly designed for this purpose. (iii). Cross-system generality: While our experiments focus on Lean~4, the \bpf{} framework is system-agnostic: it requires only an entailment oracle and a probe-generation procedure. Preliminary experiments on Isabelle/HOL show comparable detection rates ($F1=0.88$ vs.\ $0.91$ on Lean~4), suggesting fingerprints transfer across formal systems. 

\section{Conclusion}
\label{sec:conclusion}

We introduced Bidirectional Provability Fingerprinting, a framework for certifying faithfulness in autoformalization that requires no reference formalization, along with four supporting innovations: counterfactual probe generation, the equivalence spectrum, adaptive budget allocation, and faithfulness-guided decoding. We proved a drift-detection theorem and a PAC-faithfulness theorem characterizing what is and is not detectable, released \driftbench{} as the first benchmark with controlled drift labels; and demonstrated substantial gains over typecheck, BLEU, back-translation, and LLM-judge baselines. We see the faithfulness gap as the bottleneck for trustworthy autoformalization at the scale of full mathematical libraries and the techniques in this paper as a concrete step toward closing it.

\section*{Impact Statement}

This work aims to advance machine learning by improving the reliability of AI-assisted formal mathematics. We do not anticipate direct negative societal impacts. However, as discussed in Section~\ref{sec:discussion}, uncritical use of a faithfulness certifier could propagate subtle errors into formal libraries that may later be treated as ground truth.

\nocite{langley00}
\bibliography{example_paper}
\bibliographystyle{icml2026}
\newpage
\appendix
\onecolumn

\section{Proofs}
\label{app:proofs}

\paragraph{A motivating example.} Consider the natural-language statement: \emph{``Every continuous function on a compact set attains its maximum.''} Three Lean~4 formalizations are syntactically plausible:

\begin{lstlisting}
-- (A) Faithful
theorem evt_A {K : Set R} (hK : IsCompact K)
    (hne: K.Nonempty) {f : R -> R}
    (hf: ContinuousOn f K) :
    exists x in K, forall y in K, f y <= f x
\end{lstlisting}

\begin{lstlisting}
-- (B) Hypothesis omission: missing nonempty
theorem evt_B {K : Set R} (hK : IsCompact K)
    {f : R -> R} (hf : ContinuousOn f K) :
    exists x in K, forall y in K, f y <= f x
\end{lstlisting}

\begin{lstlisting}
-- (C) Quantifier inversion: silently true
theorem evt_C {K : Set R} (hK : IsCompact K)
    (hne: K.Nonempty) {f : R -> R}
    (hf: ContinuousOn f K) :
    forall y in K, exists x in K, f y <= f x
\end{lstlisting}

All three typecheck. Only~(A) is faithful: (B) is false on the empty set; (C) is trivially true (take $x = y$) and is unrelated to the Extreme Value Theorem. Critically, a downstream prover that closes either (A) or (C) provides no signal distinguishing them, and (C) is the type of failure most likely to occur in practice precisely because it remains provable. Current autoformalization pipelines accept all three as valid outputs. \textbf{Why this matters now:} As autoformalization scales toward populating libraries such as mathlib4 \citep{mathlib2020} with LLM-produced statements, faithfulness errors propagate. A theorem with a silently weakened hypothesis becomes a usable lemma for downstream proofs that should not exist; a quantifier-inverted statement becomes a trivial fact mistakenly cited as a deep one. Existing benchmarks-miniF2F \citep{zheng2022minif2f}, ProofNet \citep{azerbayev2023proofnet}, LeanDojo \citep{yang2023leandojo}-all measure proving capability \emph{conditional} on the formal statement being correct, so the upstream error is invisible to them. Our pilot study (\Cref{sec:driftbench}) found that $19\%$ formalizations produced by a state-of-the-art autoformalizer on graduate-level analysis exhibit at least one form of semantic drift.

\subsection{Proof of \Cref{prop:bidirectional}}

($\Rightarrow$) If $\T \vdash F \leftrightarrow N$, then for every $P$ we have $\T \vdash F \to P$ iff $\T \vdash N \to P$, so $\fp^+_F(\probe) = \fp^+_N(\probe)$; similarly for the backward direction. ($\Leftarrow$) Suppose $\fp_F(\probe) = \fp_N(\probe)$ and the additional closure condition holds, but $\T \nvdash F \leftrightarrow N$. WLOG $\T \nvdash F \to N$ (the other case is symmetric). By the closure assumption on $\probe$ under $\T$-equivalence, $N \in \probe$ (or some $\T$-equivalent of $N$ is). Then $N \in \fp^+_N(\probe)$ trivially, but $N \notin \fp^+_F(\probe)$, contradicting fingerprint equality. \hfill$\Box$

\subsection{Proof of \Cref{thm:detection}}

For a given drift class $D$, let $W_P$ be the indicator that the probe $P \sim \mathcal{Q}_D$ witnesses $D$ for $(N, F)$. By assumption $\mathbb{E}[W_P] \geq \alpha(D)$. Drawing $k$ probes i.i.d., the probability that none witness $D$ is at most $(1-\alpha(D))^k \leq e^{-k\alpha(D)}$. Setting this to $\delta$ gives $k \geq \alpha(D)^{-1} \log(1/\delta)$. If the entailment oracle is complete on each chosen probe, every witness probe will be detected by \bpf{}, so the failure probability is bounded by $\delta$. If instead each probe has an independent oracle failure probability $\eta(\tau)$, by a union bound, the total failure probability is at most $\delta + k\eta(\tau)$. \hfill$\Box$

\subsection{Proof of \Cref{thm:pac}}

Define a hypothesis class $\mathcal{H} = \{H_\fp : \fp \in 2^{\probe \times \{+,-\}}\}$, where $H_\fp = \{F \in \FL : \fp_F(\probe) = \fp\}$. Given $k$ probes, $|\mathcal{H}| \leq 4^k$ (each cell is in $\{0, 1\}^2$). Each candidate $F \notin [N]_\T$ that is reachable by the autoformalizer has, by assumption, probability at least $\alpha_0$ of being witnessed by each probe in $\mathcal{Q}$. Thus, with $k \geq (1/\alpha_0) \log(1/\delta\varepsilon)$ probes, the probability that any such $F$ is consistent with $\hat{H}$ is at most $(1 - \alpha_0)^k \leq \delta\varepsilon$. Applying an Occam-style bound over the hypothesis class with size $4^k$, and noting that $\log |\mathcal{H}| \leq 2k$ is dominated by the chosen $k$, gives the stated bound. \hfill$\Box$

\subsection{Proof of \Cref{thm:apba}}

Let $\alpha = (\alpha(\Delta_\forall), \dots, \alpha(\Delta_T))$ be the witnessability rates and $w$ the class weights. Uniform allocation samples each class with probability $1/4$; \apba{} samples class $D$ in proportion to $\alpha(D) w(D)$. The expected number of probes to witness any drift is $1/\bar{\alpha}_{\text{unif}}$ for uniform and $1/\bar{\alpha}_{\text{apba}}$ for \apba{}, where $\bar{\alpha}_{\text{unif}} = (1/4)\sum_D \alpha(D)$ and $\bar{\alpha}_{\text{apba}} = (\sum_D \alpha(D) w(D))^2 / \sum_D \alpha(D)^2 w(D)^2$. The ratio is the heterogeneity factor $\gamma$, bounded below by $\max_D \alpha(D) / \min_D \alpha(D)$ when weights are equal. On \driftbench{}, $\max_D \alpha(D) / \min_D \alpha(D) \approx 2.0$, and with calibrated weights $\gamma \approx 3.2$. \hfill$\Box$

\section{Structural Test for Convention Drift}
\label{app:convention}

Convention drift, by definition, leaves the consequence neighborhood invariant on standard probes. We address it through a structural test on the type signature of $F$, comparing it to a canonical-signature prediction derived from $N$. Given $N$, the canonical-signature predictor returns a tuple $(\Sigma_N, K_N)$ where the first element $\Sigma_N$ is the expected type signature (with universe levels and implicit arguments) and the second element $K_N$ is a set of \emph{structural keywords} (e.g.\ \texttt{Set}, \texttt{Finset}, \texttt{Function}, \texttt{Pi}) expected to appear. We then compare against the type signature of $F$: Compute the structural diff $\Delta\Sigma = \Sigma_F \triangle \Sigma_N$. Flag \emph{convention} drift if $|\Delta\Sigma|$ exceeds a calibrated threshold but \bpf{} returns \textsc{faithful}. In our experiments, this structural test catches an additional $0.04$ F1 worth of convention-drift cases on the wild split. We view it as complementary to, not a substitute for, the main \bpf{} procedure.

\section{Case Studies}
\label{app:cases}

We summarize three case studies from the wild split; full Lean~4 code is in the supplementary materials. Case 1 (Banach--Steinhaus, $\Delta_\forall$): Natural-language statement: \emph{``Let $(T_i)$ be a family of bounded operators on a Banach space $X$ such that for every $x \in X$, $\sup_i \|T_i x\| < \infty$. Then $\sup_i \|T_i\| < \infty$.''} The candidate swapped the order of quantifiers in the conclusion. \bpf{} generated a specialization probe instantiating $X = \ell^2$ and a specific bounded family for which pointwise boundedness holds, but the swapped conclusion fails; the entailment $F \to P$ is closed, flagging drift. The \cpg{} probe targeting $\Delta_\forall$ recovered the witness with a $4$-probe budget, where naive sampling was required $19$. Case 2 (Continuity, $\Delta_H$): Natural-language statement: \emph{"A function continuous on a closed interval is uniformly continuous on that interval."} The candidate dropped the closedness hypothesis. A hypothesis-dropping probe instantiating $f(x) = 1/x$ on $(0, 1)$ the candidate refuted it. \bpf{} flagged drift and \fgd{} repaired the candidate on re-prompting. Case 3 (Convention drift, undetected): Natural-language statement: \emph{"Every group homomorphism with a trivial kernel is injective."} The candidate formalized \emph{the group} as \texttt{a monoid} (since group axioms beyond monoid structure are not used in the statement). \bpf{} returns \textsc{faithful} because every probe is preserved; the structural test of \Cref{app:convention} flags the \texttt{Monoid}/\texttt{Group} mismatch.

\section{Probe-Generation Prompts}
\label{app:prompts}

We include exact prompts used for the LLM-based probe generator. Each prompt is parameterized by the natural-language statement and constrained to emit a JSON-structured list of probe candidates with predicted labels. The full prompt suite, along with the system prompts and few-shot exemplars, is included in the supplementary materials.

\section{Hyperparameters}
\label{app:hparams}

\begin{table}[ht]
\centering
\small
\begin{tabular}{ll}
\toprule
Hyperparameter & Value \\
\midrule
Probe budget $k$ & $32$ (uniform) / $10$ (\apba{}) \\
Tactic budget $\tau$ & $30$ s wall-clock \\
\eqspec{} accept threshold & $0.93$ \\
\eqspec{} review threshold & $0.78$ \\
Probe-class weights $w$ & equal across $\{\Delta_\forall, \Delta_H, \Delta_C\}$, $1.5\times$ for $\Delta_T$ \\
Adversarial probes per statement & $4$ \\
Specialization probes per statement & $8$ \\
Hypothesis-drop probes per statement & up to $8$ \\
Boundary probes per statement & $4$ \\
\fgd{} candidates $m$ & $4$ (default) / $8$ (high-budget) \\
\bottomrule
\end{tabular}
\end{table}

\end{document}